\documentclass[letterpaper, 10 pt, conference]{ieeeconf}
\usepackage{times}
\usepackage[pdftex]{graphicx}
\usepackage{amsmath,amssymb,amsopn,amstext,amsfonts}
\usepackage{cancel}
\usepackage[space]{cite}
\usepackage{pdfsync}
\usepackage{balance}
\usepackage{color}
\usepackage{mathtools}
\usepackage{algpseudocode}
\usepackage{algorithm}
\usepackage{algpseudocode}
\usepackage{bm}

\usepackage{float}
\usepackage{epstopdf}
\usepackage{pifont}
\usepackage{fixltx2e}
\usepackage{amsmath}
\usepackage{multirow}
\usepackage{url}
\usepackage{verbatim}
\usepackage{siunitx} 
\usepackage{epstopdf}
\usepackage{svg}
\usepackage{bm}
\usepackage{diagbox}
\usepackage{float}
\usepackage{epstopdf}
\usepackage{pifont}
\usepackage{fixltx2e}
\usepackage{multirow}
\usepackage{url}
\usepackage{verbatim}
\usepackage{siunitx} 
\usepackage{epstopdf}
\usepackage{bbding}

\usepackage{times}
\usepackage{cancel}
\usepackage{pdfsync}
\usepackage{balance}
\usepackage{color}
\usepackage{mathtools}

\usepackage[linkcolor=black,citecolor=black,urlcolor=black,colorlinks=true]{hyperref}
\usepackage{booktabs}
\usepackage{makecell}
\usepackage{enumerate}
\usepackage{arydshln}
\usepackage{etoolbox}
\makeatletter
\patchcmd{\@makecaption}
  {\scshape}
  {}
  {}
  {}
\makeatletter
\patchcmd{\@makecaption}
  {\\}
  {.\ }
  {}
  {}
\makeatother
\def\tablename{Table}

\PassOptionsToPackage{prologue,dvipsnames}{xcolor}
\makeatletter
\let\NAT@parse\undefined
\makeatother
\usepackage{subfigure}
\usepackage{caption}
\addcontentsline{toc}{section}{Acknowledgement}

\newcommand{\mbf}[1]{\mathbf{#1}}

\graphicspath{{img/}}

\DeclareGraphicsExtensions{.png,.jpg,.eps,.pdf}
\IEEEoverridecommandlockouts
\title{\LARGE \bf Star-Searcher: A Complete and Efficient Aerial System for \\ Autonomous Target Search in Complex Unknown Environments}

\author{Yiming Luo$^{1}$, Zixuan Zhuang$^{1}$, Neng Pan$^{2}$, Chen Feng$^{3}$, \\ Shaojie Shen$^{3}$, Fei Gao$^{2}$, Hui Cheng$^{1}$, Boyu Zhou$^{1, \dag}$
 	\thanks{\textsuperscript{\dag} \textbf{Corresponding Author}.}
      \thanks{\textsuperscript{1} Sun Yat-Sen University, China.}
      \thanks{\textsuperscript{2} State Key Laboratory of Industrail Control Technology, Institute of Cyber-Systems and Control, Zhejiang University, Hangzhou, China.}
	\thanks{\textsuperscript{3} Department of Electronic and Computer Engineering, The Hong Kong University of Science and Technology, Hong Kong, China.}
	\thanks{\scriptsize \{\href{mailto:yiming.luo2001@gmail.com}{yiming.luo2001@gmail.com},
         \href{mailto:zhouby23@mail.sysu.edu.cn}{zhouby23@mail.sysu.edu.cn}\}}
}

\begin{document}
\algrenewcommand\algorithmicrequire{\textbf{Input:}}
\algrenewcommand\algorithmicensure{\textbf{Output:}}

\maketitle
\thispagestyle{empty}
\pagestyle{empty}

\begin{abstract}   
This paper tackles the challenge of autonomous target search using unmanned aerial vehicles (UAVs) in complex unknown environments. To fill the gap in systematic approaches for this task, we introduce Star-Searcher, an aerial system featuring specialized sensor suites, mapping, and planning modules to optimize searching. Path planning challenges due to increased inspection requirements are addressed through a hierarchical planner with a visibility-based viewpoint clustering method. This simplifies planning by breaking it into global and local sub-problems, ensuring efficient global and local path coverage in real-time. Furthermore, our global path planning employs a history-aware mechanism to reduce motion inconsistency from frequent map changes, significantly enhancing search efficiency. We conduct comparisons with state-of-the-art methods in both simulation and the real world, demonstrating shorter flight paths, reduced time, and higher target search completeness. Our approach will be open-sourced for community benefit \footnote{\url{https://github.com/SYSU-STAR/STAR-Searcher}}.

\end{abstract}

\begin{keywords}
Aerial Systems: Perception and Autonomy, Aerial Systems: Applications, Search and Rescue Robots
\end{keywords}

\IEEEpeerreviewmaketitle

\section{Introduction}

Unmanned aerial vehicles (UAVs) are prized for their compact size and exceptional maneuverability, making them indispensable across various applications like disaster search and rescue, resource exploration, and environment monitoring. In these tasks, UAVs can effectively supplant humans in the exploration of entirely unknown and hazardous environments while simultaneously conducting target search. This paper focuses on the challenge of autonomous searching for targets in complex unknown environments using UAVs.

The challenge of autonomous target search is closely connected to the field of autonomous exploration, a fundamental domain in robotics that has garnered significant attention \cite{zhou2021fuel,cao2021tare, dang2018autonomous, kim2022autonomous, Papatheodorou_ICRA2023}. While both areas share certain similarities, they are fundamentally distinct. Autonomous exploration primarily focuses on mapping unknown regions as either occupied or free areas. {In contrast, autonomous target search demands the UAV to perform two related but different tasks simultaneously, i.e., exploration and inspection.} The former task only requires coarsely mapping the unknown space, while the latter demands meticulous visual inspections in the occupied spaces where potential targets may be located, with more rigorous constraints such as observation distance and viewing angles. Thus, an efficient system capable of handling the diverse perception requirements and generating motions that switch seamlessly between the two tasks is essential for fast target search. Currently, there is a gap in the existence of a systematic approach specifically tailored for autonomous target search—one that can ensure search completeness without compromising task efficiency.

\begin{figure}[t]
	\centering
	\includegraphics[width=0.8\linewidth]{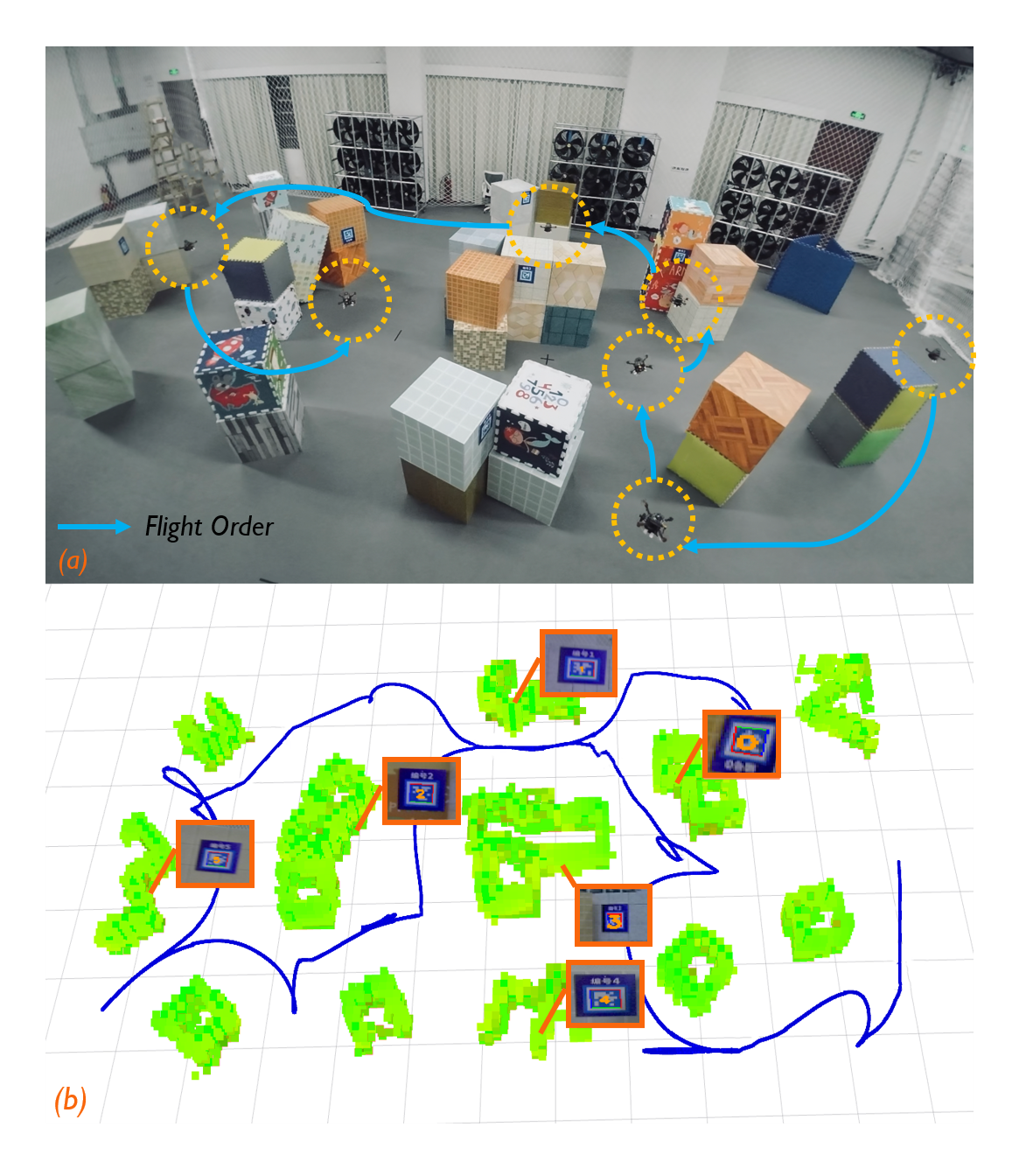}
 \vspace{-0.3cm}
	\caption{\fontsize{8.5bp}{17bp}(a) A test of autonomous target search conducted in a complex scene with six apriltags. (b) The apriltag search results and executed trajectory. Video of the experiments is available at: \href{https://youtu.be/08ll_oo_DtU}{https://youtu.be/08ll\_oo\_DtU}.}
	\label{fig_cover}
        \vspace{-1.2cm}
\end{figure}

Notably, autonomous target search presents a significant challenge in the context of path planning. The additional requirement of careful inspection introduces a substantial number of inspection viewpoints, leading to a considerably increased computational load for determining the shortest paths. Additionally, as the scene's map and uninspected areas are incrementally constructed during the search process, the UAV must adapt its path when the map undergoes changes. Such alterations in the map can cause the newly planned path to noticeably deviate from the previous one, resulting in back-and-forth movements that detrimentally impact search efficiency. For a smooth and graceful flight, it is crucial to plan paths in real-time and ensure that consecutive paths remain consistent. 

To tackle the challenges outlined above, we present \textbf{Star-Searcher}: A Complete and Efficient Aerial \textbf{S}ystem for Autonomous \textbf{Tar}get \textbf{Search} in Complex Unknown Environments. Our aerial system incorporates specialized sensor suites, mapping, and planning modules, all geared toward improving task efficiency and completeness. {Our system takes advantage of various sensors and seamlessly switches between exploring unknown space and inspecting surface regions.} We address the path planning challenges from two key angles. {Firstly, we introduce a hierarchical planner supported by a visibility-based viewpoint clustering method. This approach decomposes the complex large-scale planning task into two more manageable sub-tasks: global path planning at the level of viewpoint clusters and local path planning at the level of individual viewpoints. Viewpoint clustering groups viewpoints separated by obstacles into distinct clusters and aggregates mutually visible ones into convex sets. This strategy provides regional guidance for generating reasonable global paths that sequentially visit different regions and ensures straightforward local paths for covering each convex set. Secondly, to mitigate motion inconsistency arising from frequent map changes, our global path planning incorporates a history-aware mechanism, taking into account both the historical movement tendency and visiting costs to all viewpoints. This prevents the occurrence of indecisive global paths across consecutive planning iterations, significantly enhancing search efficiency.} 

We compared our approach to state-of-the-art fast exploration and object-centric search methods in simulations. The results demonstrated superior search performance, with the shortest path length, flight time, and the highest completeness in all experiments. We further validated our system in complex real-world environments using entirely onboard devices. We plan to release our code as open source. In summary, our contributions are as follows:
\begin{itemize}
 \item {An aerial system with specialized sensor suites, mapping, and novel planning modules, which enables seamless exploration and inspection, achieving comprehensive and fast autonomous target search in complex unknown environments.} 
 
 \item {A hierarchical planning method enhanced by visibility-based viewpoint clustering, allows the real-time generation of the global path at the level of viewpoint clusters and local path at the level of individual viewpoints with fewer detours.}

 \item {A history-aware mechanism for global path planning, which utilizes historical path information to prevent inconsistency in consecutive planning processes, significantly improving task efficiency.}

 \item Extensive simulations and real-world experiments for validation. The source code will be made public.
\end{itemize}

\section{Related Works}

\begin{figure*}[t]
	\centering
	\includegraphics[width=1\linewidth]{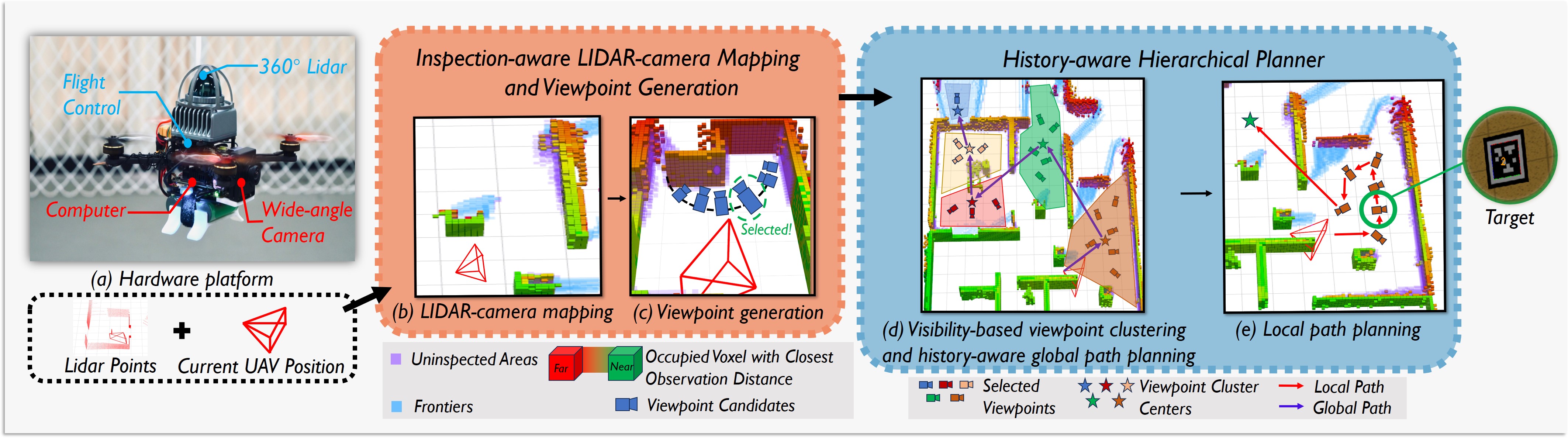}
        \vspace{-0.5cm}
	\caption{\fontsize{8.5bp}{17bp}An overview of our star-searcher system. (a) The hardware platform of our aerial system. (b) The UAV utilizes LIDAR point clouds and camera projection to update map information, cluster frontiers and uninspected areas. (c) For each cluster of frontiers and uninspected areas, a set of viewpoints is generated and scored based on information gain and viewing angle. The higher the score, the larger the size of the viewpoint. The one with the highest score is selected. (d) Visibility-based viewpoint clustering is performed, and history-aware global path planning is conducted based on this. (e) Local path planning.}
	\label{fig:overview}
    \vspace{-0.4cm}
\end{figure*}

In autonomous target search, a critical step involves exploring unknown environments. Various methods for rapid autonomous exploration have been extensively researched, among which frontier-based approaches are among the most popular methods \cite{yamauchi1997frontier, gao2018improved, faigl2013determination, kulich2019integrated, dornhege2013frontier, heng2015efficient}. The concept of the frontier is initially introduced to delineate the boundaries between unknown and known areas \cite{yamauchi1997frontier}. They employ a greedy approach that selects the nearest frontier at each step. Many subsequent methods have proposed more rational selection strategies to enhance exploration efficiency. A next-best-view selection strategy is introduced \cite{bircher2016receding} and widely used. Each sampled viewpoint is evaluated using a utility function, which assesses the information gain achievable by visiting them and the required path length. The design of the utility function has also received more extensive consideration \cite{akbari2021informed, schmid2020efficient, xu2021autonomous, gonzalez2002navigation}. Recent methods formulate the problem of traversing all viewpoints as a traveling salesman problem \cite{zhou2021fuel, cao2021tare, zhou2023racer}, improving task efficiency.

Building upon exploration planning methods, several approaches have introduced object search techniques during the exploration process \cite{dang2018autonomous, kim2022autonomous, Papatheodorou_ICRA2023, best2018online, ashour2020exploration, asgharivaskasi2021active}. Some of these methods involve conducting informative sampling to facilitate re-observations at higher resolutions when detecting objects \cite{dang2018autonomous}. Papatheodorou et al. \cite{Papatheodorou_ICRA2023} propose a utility function tailored to object-centric exploration, ensuring sufficient proximity for all background voxels. Kim et al. \cite{kim2022autonomous} apply a 2D space segmentation method and integrate search into the exploration. Meera et al. \cite{meera2019obstacle} employ a Gaussian Process-based model for target occupancy. Besides, motivated by the DARPA Subterranean Challenge, some systems have been developed to combine multiple robots to cooperate in searching objects \cite{kulkarni2022autonomous, roucek2021system, best2022resilient}. A cooperative exploration strategy is proposed to enable the robots to coordinate their exploration while having the ability to explore individually \cite{kulkarni2022autonomous}. Roucek et al. \cite{roucek2021system} develop a heterogeneous exploration robotic system and improve single-agent robustness. Other methods \cite{best2022resilient} combine multiple sensors in the task to improve the perception precision.

However, these methods trade-off between exploration and target search, lacking a more detailed consideration of autonomous target search and suffering from severe efficiency issues. In contrast, we explicitly define the problem of autonomous target search and design a suitable aerial platform. Furthermore, our history-aware hierarchical planner achieves fast search with a concise flight path.

\section{Problem Formulation}
\label{formulation_env}

The problem of autonomous target search involves the search for an unknown number of targets detectable by visual sensors within a bounded 3D space $V \subset \mathbb{R}^3$. This space is represented as a set of cubic voxels. Continuous updates of the occupancy probability $P_o(v)$ for each voxel $v$ incrementally map the initially unknown space $V_{unk} = V$ into two parts: $V_{free} \subset V$ (free space) and $V_{occ} \subset V$ (occupied space). {Since the visible surfaces of the targets, denoted as $V_{tar}$, fall within the volume of $V_{occ}$, the UAV must conduct a thorough observation of all the voxels within $V_{occ}$ to guarantee complete search. Voxels within $V_{occ}$ that have not yet been observed with the desired precision by the visual sensor are labeled as $V_{uni}$. As most sensors' perception stops at surfaces, some hollow or corner spaces can not be mapped. These spaces are denoted by $V_{res}$. The task is considered fully completed when the entire area has been searched, i.e., when$V_{free} \cup V_{occ} = V \setminus V_{res}$, and $V_{uni} = \varnothing$.} 
 \vspace{-0.4cm}
\section{System Design}
\label{system_design}

As explained in Sect. \ref{formulation_env}, autonomous target search involves the simultaneous identification of occupied and free spaces while carefully observing uninspected areas $V_{uni}$. {Improving task efficiency can be accomplished by employing various sensors to identify $V_{occ}$ and update $V_{uni}$.} We configure a UAV equipped with a 360-degree LIDAR and a wide-angle RGB camera, as shown in Fig. \ref{fig:overview}(a). Despite its conventional appearance, this platform consists of minimal sensor components carefully tailored to our specific problem. The wide sensing range of the 360-degree LIDAR facilitates the rapid acquisition of geometric information about the surrounding environment, enabling the UAV to quickly identify occupied areas for detailed inspection using the RGB camera. By utilizing the wide-angle camera, the UAV can cover more spaces and detect targets more rapidly. 

The algorithm framework in Fig. \ref{fig:overview}(b)-(e) comprises mapping and planning modules. {The mapping module represents the environment volumetrically, fusing LIDAR and camera data continuously for each voxel to update its occupancy and the closest observation distance from the camera} (Sect. \ref{mapping}). Frontiers and uninspected areas are then extracted and clustered, generating and selecting corresponding viewpoints with high information gain and suitable viewing angle (Sect. \ref{vp_generation}). The history-aware hierarchical planner (Sect. \ref{Hierarchical_planning}) then utilizes the viewpoints and previous path planning results to plan global paths and local paths, achieving simultaneous identification of free and occupied spaces and thorough inspection of uninspected areas. A visibility-based viewpoint clustering method is employed during global path planning (Sect. \ref{vp_clustering}) to create a more reasonable visiting order and reduce the computation burden.

\section{Inspection-aware LIDAR-camera Mapping and Viewpoint Generation}

Traditional occupancy mapping used in exploration lacks information about the observation distances of voxels. This limitation hinders the ability to plan inspection paths to ensure a thorough target search. Moreover, existing methods overlook the consideration of viewing angles towards uninspected surfaces, potentially leading to missed targets due to the detrimental effects of large viewing angles on target detection. Our mapping module addresses the first issue by integrating measurements from both LIDAR and the camera, providing the closest observation distance for each occupied voxel and enabling more precise path planning for thorough inspections. Additionally, we introduce a scoring mechanism for viewpoint selection, which prevents large viewing angles in uninspected areas.

\subsection{LIDAR-camera Mapping with Inspection Information}
\label{mapping}

Our volumetric environment representation is based on \cite{han2019fiesta}. In addition to occupancy probability, we update the closest observation distance to the camera for each occupied voxel. When a new frame of LIDAR point clouds is acquired, all point clouds are employed to update the occupancy probability using the ray-casting method. Simultaneously, we can project the voxels already mapped as occupied into the camera coordinate frame, updating the closest observation distances for those within the camera's field of view (FOV) and not occluded by any occupied areas. Voxels with a closest observation distance greater than the maximum observation distance $d_{max}$ are labeled as uninspected areas $V_{uni}$. They are subsequently removed from the uninspected areas when they are scanned within $d_{max}$. Continuous target detection is performed in RGB images, and when a target is detected, the corresponding voxels are mapped to $V_{tar}$ through transformation from pixel coordinates to world coordinates. It's worth noting that additional information pertaining to observation accuracy can be integrated into the map using a similar approach. However, conducting a comprehensive study of these related factors falls beyond the scope of our research.

\subsection{Viewpoint Generation}
\label{vp_generation}

 Given the information on occupancy and observation distance, both frontiers \cite{yamauchi1997frontier} and uninspected areas are extracted, as defined in Sect. \ref{formulation_env}. Uninspected areas and frontiers imply the potential presence of targets or areas that can be expanded upon in the map. Hence, we perform viewpoint sampling in proximity to these areas. Similar to \cite{zhou2021fuel}, we employ a PCA-based method to split excessively large clusters along axes. Several viewpoint positions and their corresponding yaw angles are sampled in the spherical space around the center of each cluster. Then we score all viewpoints of each cluster based on two criteria: information gain and viewing angle.

\begin{itemize}
	\item \textbf{Information Gain}: The information obtained by the UAV at each viewpoint is determined by a weighted combination of {the number of observable frontiers to the lidar $\mathcal{N}_{unknown}$ and the number of observable uninspected voxels to the camera $\mathcal{N}_{uninspected}$ within $d_{max}$, i.e.}
\begin{equation}
	\begin{aligned}
            S_{info}= \omega_{uni}\cdot\mathcal{N}_{uninspected}+ \omega_{unk}\cdot\mathcal{N}_{unknown}
	\end{aligned}
    \label{equ 1}
\end{equation}
where $ \omega_{uni}$ and $\omega_{unk}$ denote the weights for uninspected voxels and frontiers, respectively. We assign a larger value to $\omega_{uni}$ in experiments.
	\item \textbf{Viewing Angle}: To ensure accurate target detection and mitigate errors caused by extreme viewing angles, our method scores viewpoints based on the proximity between the vector from the center of the cluster to the viewpoint and the average normal vector of each cluster {${n}_{avg}$}, i.e.
\begin{equation}
	\begin{aligned}
            S_{nor}=\frac{\textbf{p}_{c,v}\cdot \textbf{n}_{avg}}{||{\textbf{p}}_{c,v}||\ ||\textbf{n}_{avg}||}
	\end{aligned}
\end{equation}
\end{itemize}

{The average normal of each cluster is computed as the average normal of the center points of each voxel within the cluster by PCL package.} The final score for each viewpoint is calculated as follows:
\begin{equation}
	\begin{aligned}
            S_{VP}= S_{nor} \cdot S_{info}
	\end{aligned}
\end{equation}
Finally, we pick the highest-scoring viewpoint of each cluster as illustrated in Fig. \ref{fig:overview}(c).

% \begin{algorithm}[t]
% \caption{Viewpoint Clustering}\label{alg:vpcluster}
% \begin{algorithmic}[1]
% \Require UAV position $P_{cur}$,  waitlist of all viewpoints $WL_{vp}$ 
% \State New set of cluster $\xrightarrow{}$ $clu\_set$ 
% \State $WL_{vp}$.add($P_{cur}$)
% \While {!$WL_{vp}$.empty()}
% \State New cluster $\xrightarrow{}$ $clu_{new}$ 
% \For{each $v_i$ in $WL_{vp}$}
% \State \textbf{if}\ Dist($clu_{new}.center$, $v_i$)$\ <R_{thresh}$ \textbf{and} 
% \State RayCast($v_i$, $v_j$)==FREE for all  $v_j$ in $clu_{new}$
% \State \hspace{2mm} $clu_{new}$.push($v_i$)
% \State \hspace{2mm} $WL_{vp}$.pop($v_i$)
% \EndFor
% \EndWhile
% \Ensure vector of all $clu_{new}$ 
% \end{algorithmic}
% \end{algorithm} 

\section{History-aware Hierarchical Planner}
\label{Hierarchical_planning}

\begin{figure}[t]
	\centering
	\includegraphics[width=0.8\linewidth]{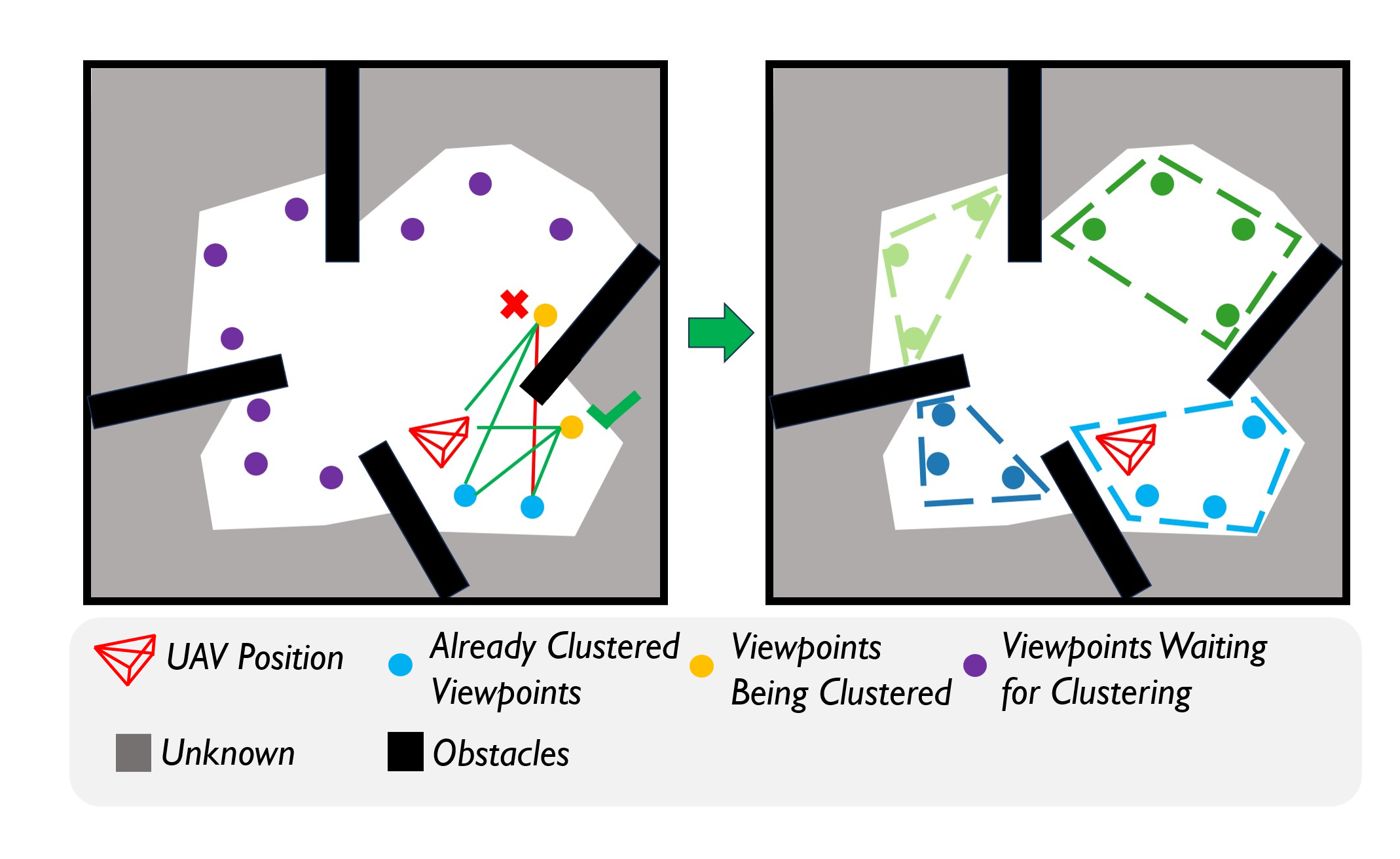}
        \vspace{-0.4cm}
	\caption{Illustration of the visibility-based viewpoint clustering. If a viewpoint has collision-free rays to all viewpoints in the current cluster, it is assigned to that cluster. }
	\label{fig:VP-cluster}
        \vspace{-1.2cm}
\end{figure}

{In order to plan paths with fewer detours and revisits in real-time, simultaneously exploring unknown regions and covering uninspected areas, we adopt a hierarchical planning strategy.} We first perform visibility-based viewpoint clustering (Sect. \ref{vp_clustering}) and conduct global path planning (Sect. \ref{CGCP}) to determine the visiting order of the viewpoint clusters. Subsequently, local path planning is conducted within the first viewpoint cluster, efficiently covering the frontiers and uninspected areas (Sect. \ref{LTP}). Our global planner incorporates historical path information, maintaining the consistency between the replanned path and the previous ones (Sect. \ref{CGCP}).

\subsection{Visibility-based Viewpoint Clustering}
\label{vp_clustering}

{By leveraging a broad sensing range, the UAV can identify occupied surfaces for thorough visual inspection rapidly.} However, this results in a significant increase in the number of required viewpoints, also for those used to cover the frontier. Planning a single shortest path to visit each of these viewpoints would result in excessive computation time, making it impossible to respond promptly to environmental changes. To address this challenge, we introduce a viewpoint clustering method to intelligently divide the entire set of viewpoints into multiple subsets. We then generate global routes passing through these subsets and local paths that visit the viewpoints within the subset sequentially, creating an efficient path within a manageable computation time.

The details of the proposed visibility-based viewpoint clustering are illustrated in Fig. \ref{fig:VP-cluster}. We initiate the process by designating the current UAV position as the starting point for the first cluster. During each clustering iteration, viewpoints within a specified radius $R_{vp}$ around the current cluster's center engage in ray-casting, prioritized based on their distances from the cluster's center. If a viewpoint's rays do not intersect with obstacles from any viewpoints in an existing cluster, that viewpoint is incorporated into the cluster, and the cluster's center is recalculated as the average position of all the viewpoints within that cluster. Once a cluster is finalized, we select the nearest unclustered viewpoint to its center to serve as the starting point for a new round of clustering. This process continues iteratively, gradually forming distinct clusters of viewpoints based on their mutual visibility and proximity until no viewpoints remain unclustered, guiding the path planning for efficient target search.

\begin{figure}[t]
	\centering
	\includegraphics[width=0.9\linewidth]{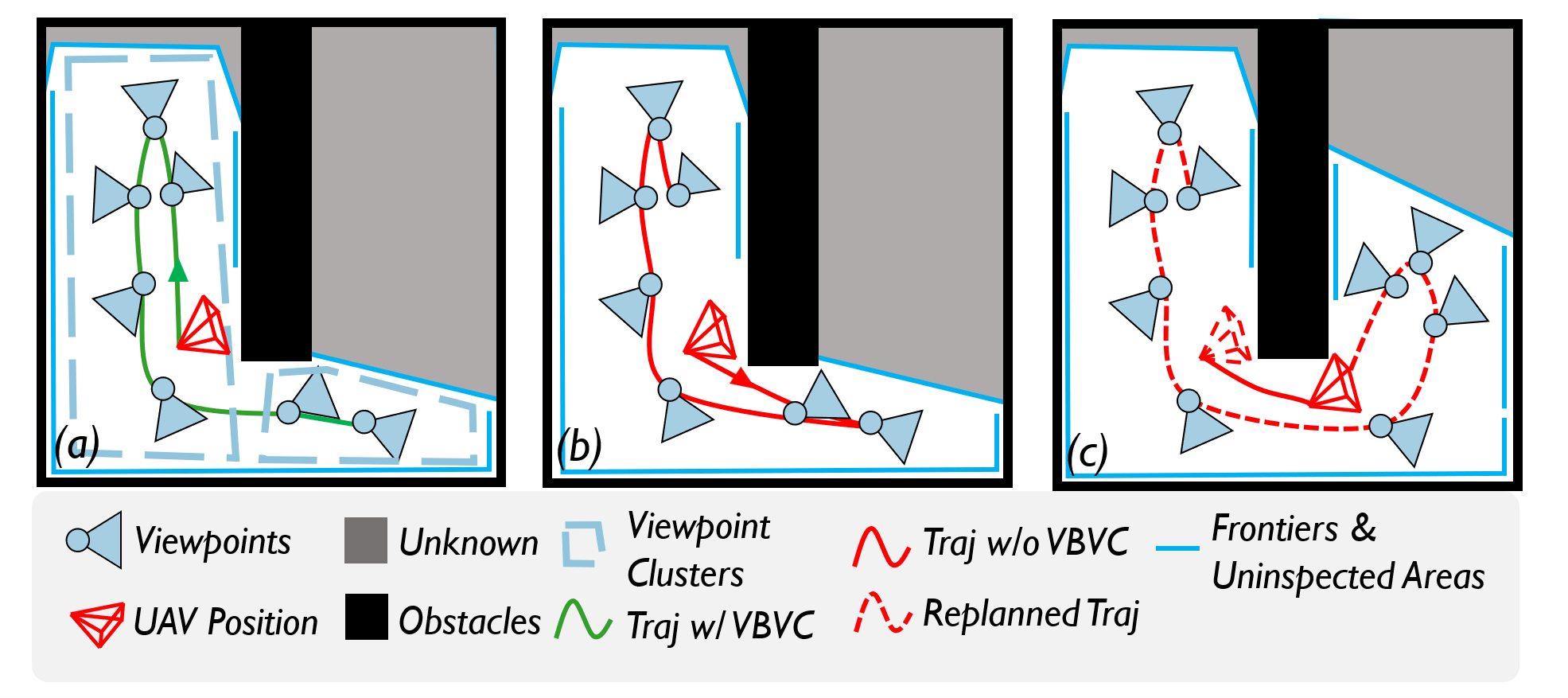}
   \vspace{-0.3cm}
	\caption{A comparison between the planned trajectories with (fig. a) and without (fig. b) visibility-based viewpoint clustering (VBVC). Without viewpoint clustering, the UAV calculates the shortest path to visit all viewpoints and selects the viewpoint behind the obstacle as the next target. Once it reaches this viewpoint, it discovers a new area and generates additional viewpoints. As a result, it replans a trajectory as indicated by the red dashed line, leading to revisits later, as shown in fig. c.}
	\label{vp-cluster-func}
        \vspace{-0.4cm}
\end{figure}

The proposed clustering method ensures mutual visibility of viewpoints within each cluster, offering several advantages. Firstly, visibility allows viewpoints in the same cluster to be encompassed by a collision-free convex set. This means the UAV can efficiently visit all viewpoints in the same cluster without the need to detour to avoid collisions. Secondly, viewpoints that are occluded by obstacles are naturally grouped into different  clusters, providing implicit regional guidance for planning a sensible global path that subsequently visits separate regions. {Additionally, this method effectively reduces severe revisits.} For instance, Fig. \ref{vp-cluster-func}(b)-(c) illustrates a scenario where the UAV prioritizes a viewpoint behind an obstacle, discovers another unexplored area, and diverts to it, leading to severe revisits later. In contrast, by thoroughly exploring one convex cluster before moving to the next, the UAV eliminates the need to circumvent obstacles and revisit previously inspected areas.

\subsection{History-aware Global Path Planning}
\label{CGCP}

\begin{figure}[]
	\centering
	\includegraphics[width=0.8\linewidth]{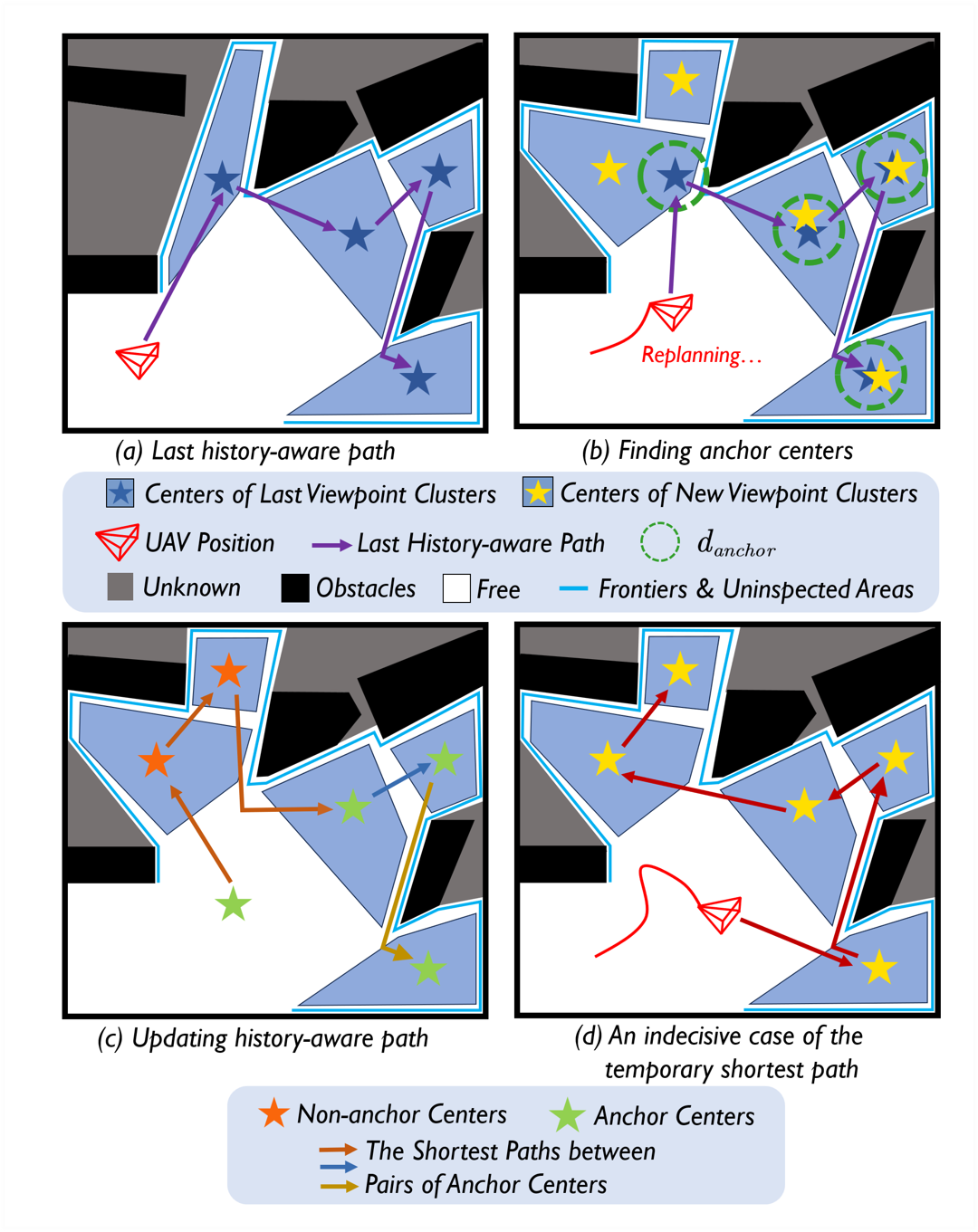}
    \vspace{-0.3cm}
	\caption{Illustration of the history-aware global path planning. After finding the anchor centers, multiple collision-free shortest paths are calculated using the A* algorithm between pairs of anchor centers and concatenated to update the history-aware global path. Without the history-aware path, an indecisive trajectory occurs, as indicated by the red curve in (d).}
	\label{fig:globaltour}
        \vspace{-1.2cm}
\end{figure}

After the viewpoint clustering, our planner computes a short global path starting from the UAV's current position, traversing the centers of all viewpoint clusters. This can be formulated as an Asymmetric Traveling Salesman Problem (ATSP)\cite{zhou2021fuel}. The planner replans every time the map is updated. However, a complete refresh of the global path after replanning can lead to drastic variations in the visiting order of different regions due to the similar path costs, causing indecisive flight as shown in Fig. \ref{fig:globaltour}(d). Actually, only a few regions are updated at a time, while other regions remain unchanged. For these unchanged regions, maintaining the visiting order from the previously planned path can ensure consistency in planning and flight. The newly updated regions should be reasonably integrated into the visitation planning for the unchanged regions. Our proposed method leverages the relative visiting order from the previous global path to generate a new global path.

{ During each global path planning, we compute a temporary shortest path constructed by considering all viewpoint cluster centers and maintain a history-aware path. Although it requires additional computation, the update of the history-aware path incurs a minimal increase in computation time.} The history-aware path is initialized using the temporary shortest path. Fig. \ref{fig:globaltour} illustrates the updating process. {To ensure the order of new viewpoint clusters aligns with the previous history-aware path, we first identify the nearest old viewpoint cluster for each new one and calculate the distances between their centers. Subsequently, the new viewpoint clusters are sorted based on the order of their nearest centers in the last history-aware path.} In regions with minimal updates, where the new viewpoint clusters exhibit slight variations, centers with a distance change less than the threshold, $d_{anchor}$, are designated as anchor centers, and their associated viewpoint clusters are labeled as unchanged. The UAV's position is also considered an anchor center. For the non-anchor viewpoint clusters, integration with anchor clusters is required to form a coherent new path. To achieve this, we select pairs of neighboring anchor clusters as start and end points, retrieve non-anchor clusters between them, and compute the shortest path. The associated problem is formulated as a TSP with fixed start and end nodes, which can be transformed into an ATSP\cite{zhou2023racer}. Multiple such shortest paths between pairs of anchor clusters are concatenated to obtain the new history-aware path. Finally, the cost of the history-aware path is compared to the cost of the temporary shortest path. If the difference $\mathbb{D}_{cost}$ between them is small, the history-aware path is adopted. Conversely, the temporary shortest path is adopted, and the history-aware path is reset to the temporary shortest path.

\subsection{Local Path Planning}
\label{LTP}

The global path provides a rational order for visiting different regions. We perform more detailed local path planning based on the guidance of the global path. Specifically, we select all viewpoints in the first viewpoint cluster to visit for local path planning. We construct an ATSP with the UAV's current position as the starting point and the center of the second viewpoint cluster in the global path as the endpoint. To enhance the smoothness of UAV motion, we take into account the variations in the UAV's velocity when calculating the visiting cost. We determine the cost associated with moving the UAV from its current position to each viewpoint by breaking down the UAV's velocity into components aligned with and perpendicular to the line connecting the UAV to the viewpoint, modeling the motion in both directions as constant acceleration motion with an initial position at the UAV's current position, final position at each viewpoint, initial velocity as $v_{ali}$ and $v_{per}$, { final displacement as the distance from the current position to the viewpoint $l$ and 0, and calculate the corresponding cost by computing the time required for the motion in both directions. Consequently, the cost function from the UAV position $p$ to the viewpoint $vp_i$ can be defined as follows:
\begin{equation}
	\begin{aligned}
            t_{ali} =
            \begin{cases}
            \frac{\sqrt{v_{ali}^2+2a_{ali}l}-v_{ali}}{a_{ali}}, & \rm{if} \frac{v_{max}^2-v_{ali}^2}{2a_{ali}}<l \\
            \frac{v_{max}-v_{ali}}{a_{ali}}+\frac{l- \frac{v_{max}^2-v_{ali}^2}{2a_{ali}}}{v_{max}}, & \rm{else}
            \end{cases}
	\end{aligned}
\end{equation}
}
\begin{equation}
	\begin{aligned}
            \mathbb{C}(p, vp_i)=\rm{max}\bigg\{t_{ali}, \frac{2v_{per}}{a_{per}}, 
            \frac{|\phi-\phi_j|}{\omega_{max}}\bigg\}
	\end{aligned}
\end{equation}
$\phi$ and $\omega_{max}$ denote the yaw angle and the max angular velocity of the UAV respectively. The cost between two points $vp_i$ and $vp_j$ is defined as follows:
\begin{equation}
	\begin{aligned}
            \mathbb{C}(vp_i, vp_j)=max\bigg\{\frac{L(vp_i,vp_j)}{v_{max}}, 
            \frac{|\phi_i-\phi_j|}{\omega_{max}}\bigg\}
	\end{aligned}
\end{equation}
{where $L(vp_i,vp_j)$ denotes the distance between two viewpoints.} Once we obtain the local path, we follow the framework proposed in \cite{zhou2019robust} to generate a smooth and safe trajectory, which is then sent to the controller for execution.
\vspace{-0.7cm}
\subsection{Computation Complexity Analysis}
Due to the large number of generated viewpoints, attempting to use all of them for path planning would impose a significant computation burden. Within these extensive computation needs, the computation of the visiting cost between different viewpoints predominantly consumes a significant portion of resources. The farther apart two viewpoints are, the longer it takes to search for a path and calculate the visiting cost between them. However, the planned path often avoids consecutively visiting two points with a high visiting cost, making the computation unnecessary yet time-consuming. Our hierarchical planner only requires computing costs between the centers of viewpoint clusters during the global path planning and between the viewpoints within the first cluster to visit during the local path planning. Therefore, our method transforms the computation complexity from $O(N^2)$ to $O(n_1^2)$ for global path planning and $O(n_2^2)$ for local path planning. Here, $N$ represents the total number of all viewpoints, $n_1$ is the number of viewpoint clusters, and $n_2$ is the number of viewpoints within the first cluster. Typically, $N>>n_1, n_2$. Furthermore, the viewpoints in local path planning are typically in proximity, avoiding the need to calculate costs between distant viewpoints. {Consequently, our planner, which can run at a frequency of 10 Hz in both simulation and real-world experiments, achieves good real-time capability.}

\section{RESULTS}
\begin{figure*}[t]
	\centering
	\includegraphics[width=0.9\linewidth]{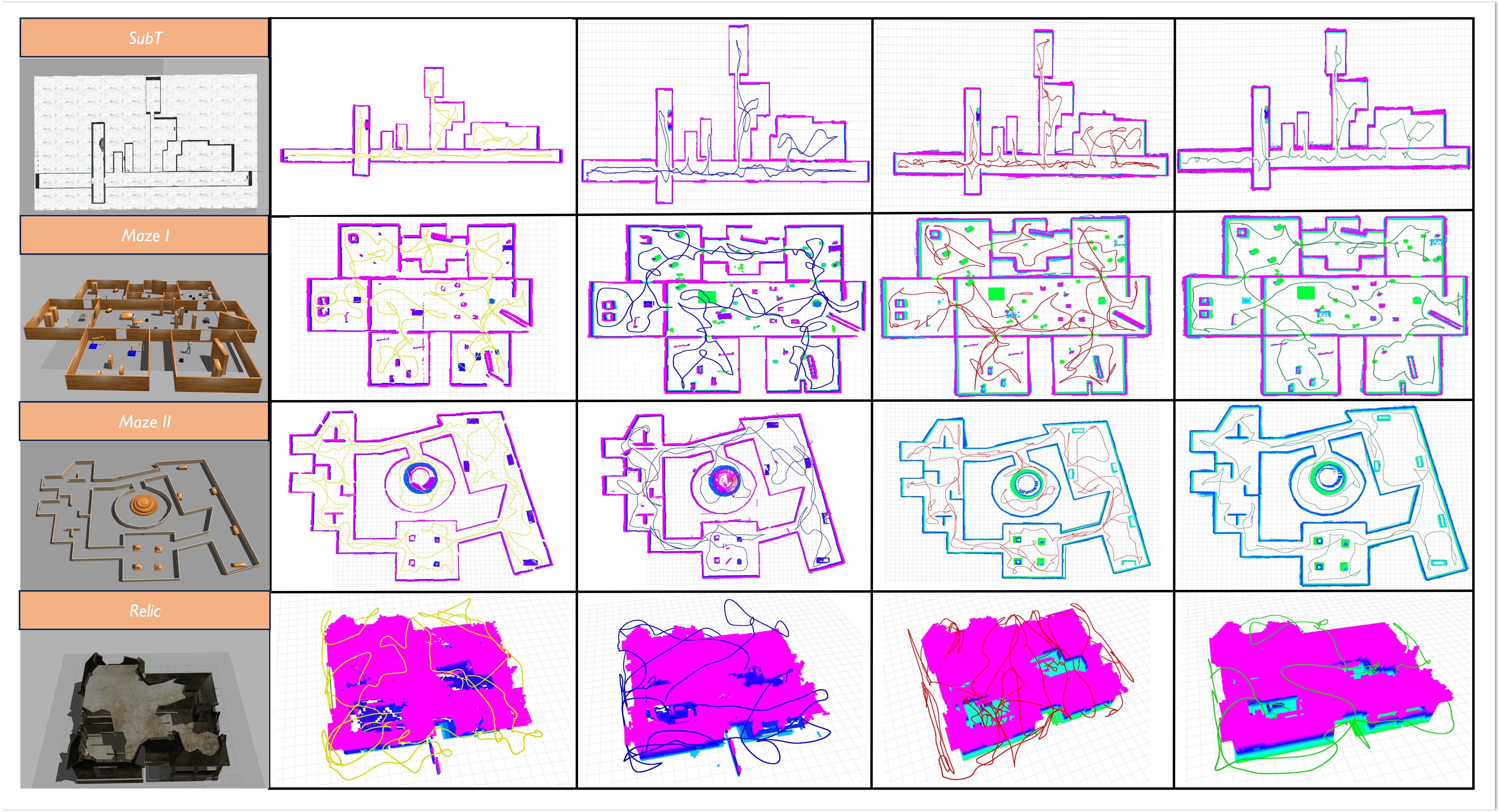}
	\caption{Trajectories generated by the proposed method (\textcolor{green!70!black}{green}), FUEL-3m\cite{zhou2021fuel}(\textcolor{yellow}{yellow}), FUEL-4m\cite{zhou2021fuel}(\textcolor{blue!70!black}{blue}) and Semantic \cite{Papatheodorou_ICRA2023}(\textcolor{red}{red}) in simulation experiments.}
	\label{simulation_result}
        \vspace{-0.8cm}
\end{figure*}

\begin{table}
 \caption{Simulation experiments.}
   \label{tab:simulation}
\centering
\resizebox{0.5\textwidth}{!}{\begin{tabular}{ccccccc} 
\hline\hline
\multirow{2}{*}{\textbf{Scene}}                                               & \multirow{2}{*}{\textbf{Method}} & \multicolumn{2}{l}{\textbf{Flight time(s)}} & \multicolumn{2}{l}{\begin{tabular}[c]{@{}l@{}}\textbf{Path} \\\textbf{length(m)}\end{tabular}} & \begin{tabular}[c]{@{}l@{}} \textbf{Complet-} \\ \textbf{eness(\%)} \end{tabular}  \\ 
\cline{3-7}
                                                                     &                         & \textbf{Avg}    & \textbf{Std}                      & \textbf{Avg}     & \textbf{Std}                                                                       & \textbf{Avg}                                                         \\ 
\cline{3-7}
\multirow{4}{*}{SubT}                                                & FUEL-3m\cite{zhou2021fuel}             & 195.19 & 11.66                      & 265.75 & 16.12                                                                   & 100                                                    \\
& FUEL-4m\cite{zhou2021fuel}                    & 177.09 & 4.17                      & 245.63 & 6.92                                                                    &\quad95.0\XSolidBrush                                                      \\
                                                                     & Semantic\cite{Papatheodorou_ICRA2023}                & 266.19 & 23.04                     & 309.70  & 29.76                                                                   & 100                                                          \\
                                                                     & Ours                    & \textbf{153.36} & 3.46                      & \textbf{191.65} & 5.31                                                                    & 100                                                          \\ 
\hline
\multirow{4}{*}{\begin{tabular}[c]{@{}l@{}}Maze I\end{tabular}} & FUEL-3m\cite{zhou2021fuel}                  & 256.48 & 5.62                      & 344.14 & 4.60                                                                  & 100                                                        \\
& FUEL-4m\cite{zhou2021fuel}                  & 236.99 & 7.58                      & 328.37 & 16.33                                                                   &\quad77.5\XSolidBrush  \\
                                                                     & Semantic\cite{Papatheodorou_ICRA2023}                 & 387.75 & 27.39                     & 426.40  & 35.93                                                                   & 100                                                          \\
                                                                     & Ours                    & \textbf{216.94} & 6.04                      & \textbf{281.69} & 14.12                                                                   & 100                                                          \\ 
\hline
\multirow{4}{*}{\begin{tabular}[c]{@{}l@{}}Maze II\end{tabular}} & FUEL-3m\cite{zhou2021fuel}                   & 394.58 & 31.30                    & 559.96 & 36.26                                                                 & 100                                                        \\
& FUEL-4m\cite{zhou2021fuel}                   & 308.01 & 40.36                     & 468.71 & 35.89                                                                    &\quad68.8\XSolidBrush                                                       \\
                                                                     & Semantic\cite{Papatheodorou_ICRA2023}                 & 502.7  & 29.09                    & 678.50  & 54.73                                                                   & 100                                                          \\
                                                                     & Ours                    & \textbf{303.15}  & 21.78                      & \textbf{458.83} & 26.62                                                                & 100                                                          \\
                                                                     \hline
\multirow{4}{*}{\begin{tabular}[c]{@{}l@{}}Relic\end{tabular}} & FUEL-3m\cite{zhou2021fuel}                   & 261.68 & 6.42                 & 371.41 & 15.70                                                                  & 100                                                      \\
& FUEL-4m\cite{zhou2021fuel}                   & \textbf{178.87} & 13.61                    & 283.57 & 17.51                                                                  &\quad81.3\XSolidBrush                                                       \\
                                                                     & Semantic\cite{Papatheodorou_ICRA2023}                 & 368.77  & 19.66                 & 563.66  & 30.18                                                                  & 100                                                          \\
                                                                     & Ours                    &  179.33 & 9.67                  & \textbf{269.64} & 9.59                                                                    & 100                                                                   \\
\hline\hline
\end{tabular}}
 \vspace{-0.8cm}
\end{table}

\subsection{Implementation Details}

We set $\omega_{uni}$ = 0.8 and $\omega_{unk}$ = 0.2 in Equ.\ref{equ 1}. In viewpoint clustering, we set $R_{vp}$ to 3m. {The ATSPs are solved using the Lin-Kernighan-Helsgaun heuristic solver\cite{helsgaun2000effective}.} For real-world experiments, we utilize the Mid360 LIDAR and an efficient LIDAR-inertial localization system \cite{xu2021fast}. A geometric controller \cite{lee2010geometric} is employed to track the $(x,y,z,\phi)$ trajectory. All modules run on the Jetson Orin NX 16GB platform.
\vspace{-0.2cm}
\subsection{Benchmark Comparisons}

We conduct simulation experiments in Gazebo, evaluating our method in four scenes: SubT \cite{best2022resilient} (68 m x 18 m x 2 m), maze I (33 m x 27 m x 2 m), maze II (60 m x 50 m x 2 m) and relic (10 m x 10 m x 7 m). We place 8 apriltags \cite{Wang2016} in each scene, with an effective recognition distance set to 3.0 m. Our proposed method is compared to FUEL \cite{zhou2021fuel} (a fast exploration method) and Semantic \cite{Papatheodorou_ICRA2023} (a NBVP-based object search method). Since Semantic \cite{Papatheodorou_ICRA2023} has not been fully open-source, we use our implementation (excluding object mapping). The proposed method and Semantic \cite{Papatheodorou_ICRA2023} employ a UAV equipped with a LIDAR with a range of 8m and an RGB camera with a fov of [68, 51] degrees. The sensor used for FUEL \cite{zhou2021fuel} is a depth camera and we conduct two experiments with different sensing ranges for FUEL \cite{zhou2021fuel} to show the limitations of applying exploration methods directly to target search. In one experiment, the sensing range is equivalent to the effective recognition distance to ensure completeness while the other one is set to 4 m. Additionally, we configure FUEL \cite{zhou2021fuel} with the same sensor setting and observation distance constraint as the proposed method in the ablation study. The side length of each voxel is set to 0.1 m, and the maximum observation distance $d_{max}$ is set to 3.0 m. We set the dynamics limits as $\mbf{\upsilon}_{\max}$ = 2.0 m/s, $\mbf{a}_{\max}$ = 1.5 m/${s^2}$ and $\mbf{\omega}_{\max}$ = 1.2 rad/s. All methods are run with the same configuration for 5 times in each scene. We evaluate the efficiency (flight length and time) and search completeness (number of recognized apriltags) of each method. Table.\ref{tab:simulation} presents the results for each scene, while Fig. \ref{simulation_result} displays the executed trajectories in different scenes.

% Considering that proposed method requires scanning all surfaces in the scene with the camera, FUEL serves as a benchmark for the proposed method in terms of target search rate and the efficiency of the sensor setup.

The results show that many revisits and detours occur in the FUEL \cite{zhou2021fuel} and Semantic \cite{Papatheodorou_ICRA2023} due to the lack of effective planning methods for target search. In contrast, our method achieves high search completeness due to ensuring sufficient observation of each occupied voxel. Also, the proposed method demonstrates the shortest path length and flight time in this task. This is attributed to the viewpoint clustering method, which prevents detours and revisits. The history-aware global path planning also produces a more consistent path, enhancing efficiency.

\vspace{-0.2cm}
\subsection{Ablation Study}
\vspace{-0.2cm}
\begin{figure}[h]
	\centering
	\includegraphics[width=1\linewidth]{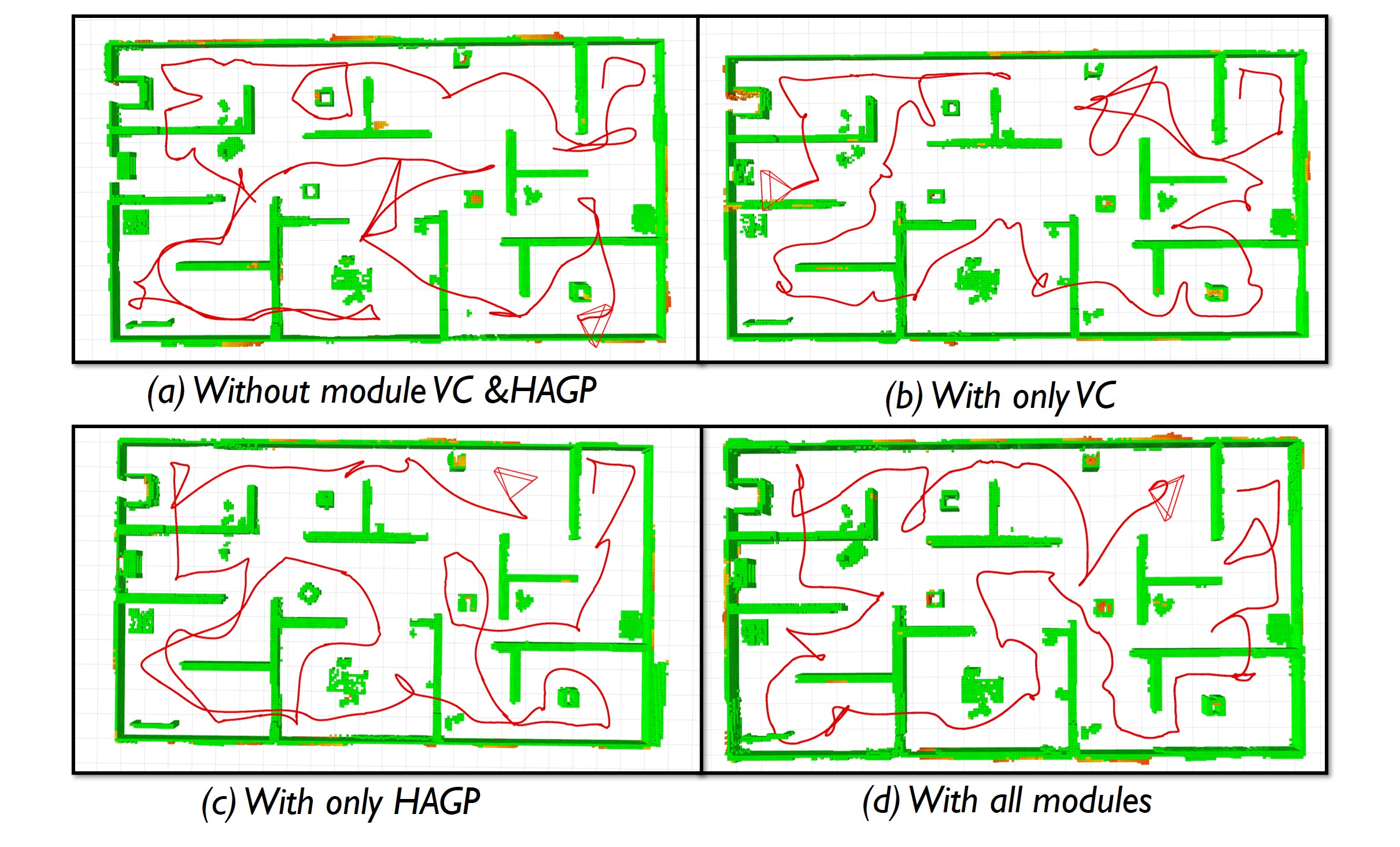}
	\caption{Executed trajectories in the ablation study. We test our method without viewpoint clustering (module VC) or history-aware global planning (module HAGP).}
    \vspace{-0.4cm}
    \label{ablation_figure}
\end{figure}

Table.\ref{ablation_tab} and Fig. \ref{ablation_figure} illustrate our tests regarding viewpoint clustering (Module VC) and history-aware global planning (Module HAGP) in the small maze (24 m x 12 m x 2 m). When neither of the two modules is engaged, it is equivalent to FUEL\cite{zhou2021fuel} with the LIDAR-camera mapping module, solving the path by considering all viewpoints. Fig. \ref{ablation_figure}(b) and the corresponding table data show that our viewpoint clustering module considers the visibility between viewpoints, reducing the detours across obstacles. Without the viewpoint clustering module, the computation time sharply increases as the number of viewpoints accumulates, causing lag in algorithm execution. {The comparison between Exp 2 and Exp 4 demonstrates the effectiveness of history-aware global planning for improving planning consistency and efficiency.} In conclusion, with all modules combined, our algorithm generates a concise and consistent path with fewer revisits, and it also exhibits good real-time capability.
 \vspace{-0.4cm}
{
\begin{table}
	\caption{Results in the ablation study.}
	  \label{ablation_tab}
   \centering
   \resizebox{0.35\textwidth}{!}{\begin{tabular}{ccccccc} 
   \hline\hline
   \multirow{2}{*}{\textbf{Setting}}                                      & \multicolumn{2}{l}{\textbf{Flight time(s)}} & \multicolumn{2}{l}{\begin{tabular}[c]{@{}l@{}}\textbf{Path} \\\textbf{length(m)}\end{tabular}}  \\ 
   \cline{2-5}
																	                       & \textbf{Avg}    & \textbf{Std}                      & \textbf{Avg}     & \textbf{Std}                                                                           \\ 
                                        \\
   \cline{2-5}
   \multirow{2}{*}{\begin{tabular}[c]{@{}l@{}}\ \ \ Exp 1 ( Without\\module VC \& HAGP)\end{tabular}}   &\multirow{2}{*}{\centering 126.25} &\multirow{2}{*}{\centering 9.89} &\multirow{2}{*}{\centering 146.15} &\multirow{2}{*}{\centering 4.46} \\
   \\
   \hline
   \multirow{2}{*}{\begin{tabular}[c]{@{}l@{}} Exp 2 (Only VC) \end{tabular}}              &\multirow{2}{*}{\centering 109.04} &\multirow{2}{*}{\centering 8.24} &\multirow{2}{*}{\centering 116.68} &\multirow{2}{*}{\centering 10.81} \\
   \\
   \hline
   \multirow{2}{*}{\begin{tabular}[c]{@{}l@{}} Exp 3 ( Only HAGP)\end{tabular}}               &\multirow{2}{*}{\centering 129.47} &\multirow{2}{*}{\centering 16.65} &\multirow{2}{*}{\centering 143.17} &\multirow{2}{*}{\centering 9.21} \\
   \\
   \hline	
   \multirow{2}{*}{\begin{tabular}[c]{@{}l@{}} Exp 4 ( With all \\ modules)\end{tabular}}           &\multirow{2}{*}{\centering 97.8} &\multirow{2}{*}{\centering 4.90} &\multirow{2}{*}{\centering 109.4} &\multirow{2}{*}{\centering 7.77} \\
 \\
 \hline\hline
   \end{tabular}}
    \vspace{-0.8cm}
\end{table}
}
\subsection{Real-world Experiments}
We further validate our method in real-world experiments. Considering camera motion blur, we set the dynamics limits as
$\mbf{\upsilon}_{\max}$ = 1.6 m/s, $\mbf{a}_{\max}$ = 0.8 m/${s^2}$ and $\mbf{\omega}_{\max}$ = 0.9 rad/s. The maximum observation distance $d_{max}$ is set to 2.0 m, within which the apriltag can be stably recognized. All of our modules run onboard without relying on external devices.

\begin{figure}[h]
	\centering
	\includegraphics[width=0.65\linewidth]{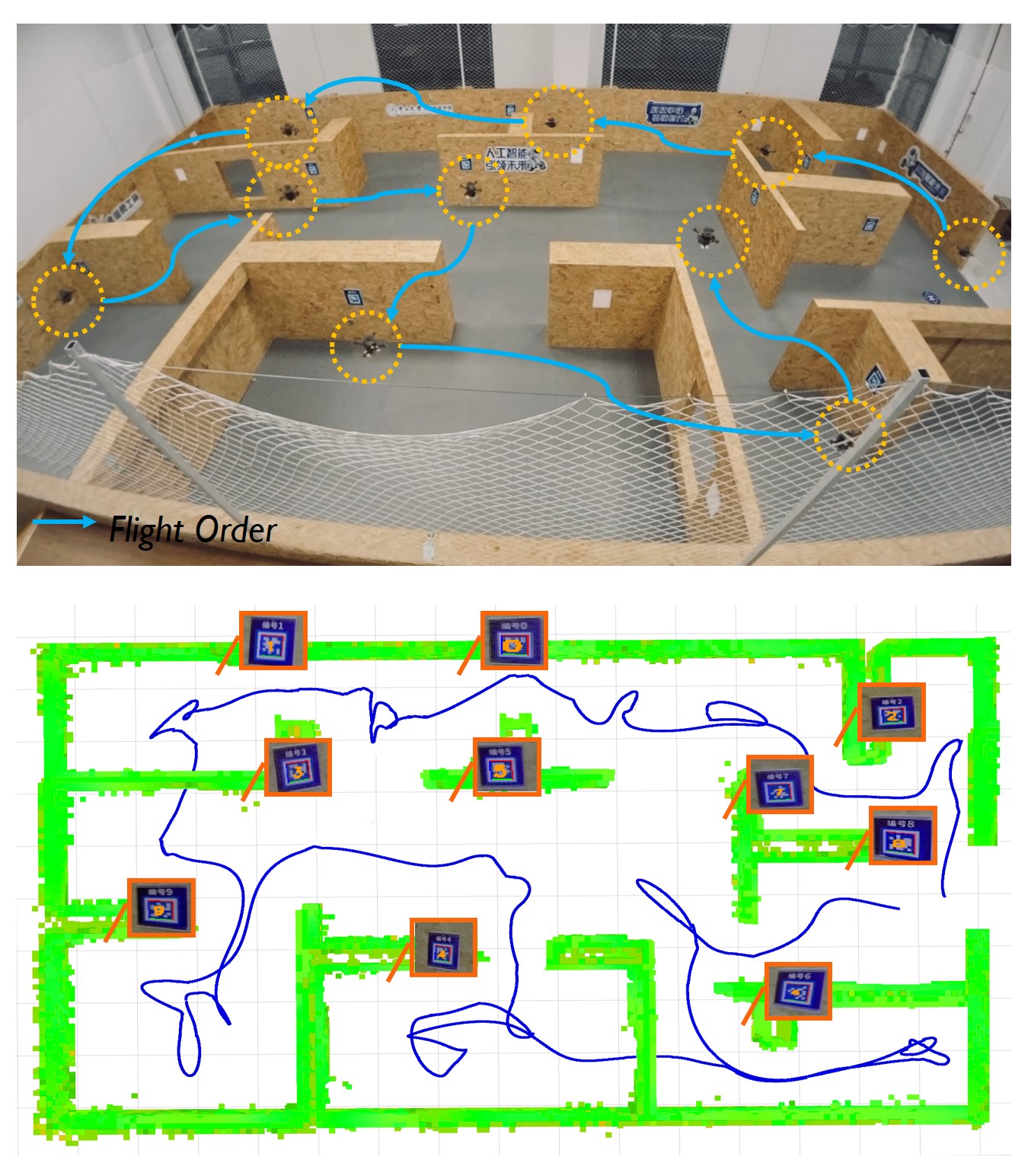}
        \vspace{-0.1cm}
	\caption{Experiments conducted in a maze with 10 apriltags to be found.}
	\label{real_result}
  \vspace{-0.8cm}
\end{figure}

The first scene consists of a cluttered indoor environment (14 m x 7 m x 2 m ) with 6 apriltags. The second scene is a maze (15 m x 8.5 m x 2 m) with 10 apriltags. Each apriltag has dimensions of 0.12 m x 0.12 m. The three-dimensional coordinates of each apriltag in the world frame are pre-measured and calibrated using laser ranging. Utilizing the UAV's pose obtained by the SLAM module \cite{xu2021fast} and the pose of the apriltags recognized in the camera frame, we could compute the apriltags' positions in the world frame. In each experiment, all apriltags are successfully recognized, with a coordinate error ranging from 0.2 m to 0.4 m. The time taken for the two scenes are 110 s and 180 s respectively, more details can be seen in the video. The online generated map and trajectory are shown in Fig. \ref{fig_cover} and Fig. \ref{real_result}. These experiments validate the capability of our system in complex real-world scenarios. 
\vspace{-0.2cm}

\section{Conclusion}
In this paper, we propose a systematic solution designed to autonomous target search in complex unknown environments. An aerial system with specialized sensor suites, mapping, and planning modules is developed to enhance task efficiency and completeness. A hierarchical planner generates global and local paths with regional guidance provided by a visibility-based viewpoint clustering method in real-time. A history-aware mechanism is introduced to prevent motion inconsistency in consecutive global planning processes. Extensive simulation and real-world experiments validate the effectiveness of our proposed method.

We have identified several directions for future work: extending the current method to a multi-UAV swarm, considering conducting autonomous target search in complex dynamic environments, and for dynamic targets.

\newlength{\bibitemsep}\setlength{\bibitemsep}{-0.08\baselineskip}
\newlength{\bibparskip}\setlength{\bibparskip}{-0.6pt}
\let\oldthebibliography\thebibliography
\renewcommand\thebibliography[1]{%
        \oldthebibliography{#1}%
        \setlength{\parskip}{\bibitemsep}%
        \setlength{\itemsep}{\bibparskip}%
}

\bibliography{RAL2023}

\end{document}